\RequirePackage{amsmath}
\documentclass[runningheads]{llncs}

\usepackage{amssymb}
\usepackage{amsmath}
\usepackage{graphicx}
\usepackage{url}
\usepackage{color}
\newcommand{\citep}[1]{\cite{#1}}
\newcommand{\citet}[1]{\cite{#1}}
\usepackage{subfig}
\usepackage{paralist}

\usepackage{fancyvrb}
\usepackage{comment}

\newcommand{\cL}{\mathcal{L}}

\newcommand{\cH}{\mathcal{H}}
\newcommand{\cF}{\mathcal{F}}

\newcommand{\cD}{\mathcal{D}}

\newcommand{\br}{\boldsymbol{r}}
\newcommand{\bd}{\boldsymbol{d}}

\definecolor{red}{rgb}{0.894, 0.102, 0.11}
\definecolor{blue}{rgb}{0.216, 0.494, 0.722}
\definecolor{green}{rgb}{0.302, 0.686, 0.29}
\definecolor{purple}{rgb}{ 0.596, 0.306, 0.639}
\definecolor{orange}{rgb}{ 1.00, 0.498, 0}
\definecolor{gray}{rgb}{0.60, 0.60, 0.60}
\definecolor{brown}{rgb}{0.651, 0.337, 0.157}
\definecolor{pink}{rgb}{0.969, 0.506, 0.749}

\newcommand{\todo}[1]{{\leavevmode\color{red}#1}}
\renewcommand{\todo}[1]{}
\newcommand{\com}[1]{{\leavevmode\color{blue}#1}}
\renewcommand{\com}[1]{}
\newcommand{\raus}[1]{{\leavevmode\color{gray}#1}}
\renewcommand{\raus}[1]{}

\begin{document}

\title{Learning Structured Declarative Rule Sets --- \\ A Challenge for Deep Discrete Learning}
\titlerunning{Learning Structured Declarative Rule Sets}

\author{Johannes F\"urnkranz\inst{1} \and Eyke H\"ullermeier\inst{2}
\and \\ Eneldo Loza Menc\'ia\inst{3} \and Michael Rapp\inst{3}}
\authorrunning{J. F\"urnkranz et al.}

\institute{Computational Data Analytics, JKU Linz, Austria, 
\and
Heinz Nixdorf Institute, Paderborn University, Germany
\and Knowledge Engineering Group, TU Darmstadt, Germany
}

\maketitle

\begin{abstract}
Arguably the key reason for the success of deep neural networks is their ability to autonomously form non-linear combinations of the input features, which can be used in subsequent layers of the network. The analogon to this capability in inductive rule learning is to learn a structured rule base, where the inputs are combined to learn new auxiliary concepts, which can then be used as inputs by subsequent rules. Yet, research on rule learning algorithms that have such capabilities is still in their infancy, which is\,---\,we would argue\,---\,one of the key impediments to substantial progress in this field. In this position paper, we want to draw attention to this unsolved problem, with a particular focus on previous work in predicate invention and multi-label rule learning.
\end{abstract}

\section{Introduction}

Current rule learning algorithms typically learn a flat set of rules, where each rule makes a prediction for a part of the instance space. These predictions are then combined via some aggregation procedure,
typically making use of a rule weight $w_i$ (e.g., corresponding to a confidence estimate of the rule), and a default rule  $\bd$, which does not need to have a weight and only makes a prediction if none of the other rules fires. Note that the use of weights and default rules goes beyond a purely logical interpretation of the underlying formula.

As an example, consider the 
rule set shown in Figure~\ref{fig:rs}.
\begin{figure}[b!]
\begin{equation*}
\begin{array}{lrlr}
\hline
\br_1(0.8): & a \wedge b & \rightarrow x  \\
\br_2(0.9): & b \wedge c & \rightarrow y \\
\br_3(0.7): & c \wedge d & \rightarrow x \\
\bd:   & & \rightarrow z & \\
\hline
\end{array}
\end{equation*}
\caption{Rule set consisting of three rules with weights and a default rule.}
\label{fig:rs}
\end{figure}
%
%
Typical aggregation procedures are, e.g., using the majority vote of the rules or the maximum weight of the rules. In our example, these two would differ for the example $(a,b,c,d)$, for which the maximum weight aggregation predicts $y$ on the basis of $\br_2$, whereas voting would predict $x$ on the basis of $\{\br_1,\br_3\}$.

A frequently used special case are so-called decision lists \citep{DecisionLists}, which organize the rules in a sequence, and the first rule that fires on an example makes the prediction. Very often, this sequence corresponds to the order in which the rules have been learned, e.g., using the separate-and-conquer or covering strategy \cite{jf:AI-Review}. Note, however, that the maximum aggregation mentioned above essentially corresponds to a decision list, in which the rules are sorted by their corresponding weights. Thus, our example would correspond to a decision list $\cD = (\br_2,\br_1,\br_3,\bd)$.

In both cases, a voted rule set or a decision list, the semantics of the rule set is not equivalent to its purely logical interpretation. However, it can be transferred to a purely declarative rule set by associating each rule with a unique hidden feature, which can be used to explicitly encode the decision procedure. 
This results in a structured rule set, in which, in addition to the input features $\cF$  (in our example $\{a,b,c,d\}$) and the output labels $\cL$  (in our example $\{x,y,z\}\}$), the rules can make use of arbitrary hidden features $\cH = \{h_i | i \in \mathbb{N}\}$. For example,
the above-mentioned weighted voting rule set and decision list could be equivalently represented as  purely declarative, structured rule sets as show in Figure~\ref{fig:structured-rs}.

\begin{figure}[h!]
\begin{minipage}{0.4\textwidth}
\begin{equation*}
\begin{array}{rlr}
\hline
a \wedge b & \rightarrow h_1  \\
b \wedge c & \rightarrow h_2 \\
c \wedge d & \rightarrow h_3 \\
h_1 \wedge h_3 & \rightarrow x \\
h_1 \wedge \neg h_2 & \rightarrow x \\
h_3 \wedge \neg h_2 & \rightarrow x \\
h_2 \wedge \neg h_1 & \rightarrow y \\
h_2 \wedge \neg h_3 & \rightarrow y \\
\neg h_1 \wedge \neg h_2 \wedge \neg h_3 & \rightarrow z & \\
\hline
\end{array}
\end{equation*}
\end{minipage}
~
\begin{minipage}{0.4\textwidth}
\begin{equation*}
\begin{array}{rlr}
\hline
b \wedge c & \rightarrow h_2  \\
h_2 & \rightarrow y \\
\neg h_2 \wedge a \wedge b & \rightarrow h_1 \\
h_1 & \rightarrow x \\
\neg h_1 \wedge \neg h_2 \wedge c \wedge d & \rightarrow h_3 \\
h_3 & \rightarrow x \\
\neg h_1 \wedge \neg h_2 \wedge \neg h_3 & \rightarrow z & \\
\hline
\end{array}
\end{equation*}
\end{minipage}
\caption{Declarative rule sets equivalent to the rule sets of Figure~\ref{fig:rs} interpreted with weighted voting (left) and as a decision list (right).}
\label{fig:structured-rs}
\end{figure}

No state-of-the-art rule learning algorithm is able to learn such structured purely declarative rule sets. For doing so, at least two key ingredients are necessary:
\begin{enumerate}
	\item Rules must be organized and learned in a way so that the predictions produced by some rules can be used by subsequent rules.
	\item A procedure must be defined for evaluating rules that have heads $h_i$, which are not defined in the training data.
\end{enumerate}
However, both ingredients can be found in multi-layer neural networks, where 1.\ corresponds to the layered architecture of the network, and 2.\ is solved via the backpropagation training protocol. 

In the following, we will briefly summarize two lines of work where some results on these problems are available: multi-label rule learning, and predicate invention.

\section{Multi-label Rule Learning}

In multi-label classification, the task is to assign a subset of the possible labels $\cL$ to each instance \citep{tsoumakas2010,zhang2013}. A standard algorithm is binary relevance (BR), which learns a separate binary model for each of the labels. Note that, when no negations can appear in the head, the resulting rule set is already purely declarative, because no additional means are necessary for making a prediction: if multiple rules make predictions for different labels, no conflict arises but all predicted labels are set, whereas if no rule covers an example, it is also fine to predict an empty set of labels (and therefore no default rule is necessary).

However, the key disadvantage of this approach is that it treats each label independently. In fact, most of the research in multi-label classification aims for the development of methods that are capable of modeling such label dependencies \citep{dembczynski2012a}. One of the best-known approaches are so-called classifier chains (CC) \citep{read2009,cheng2010}, which learn the labels in some (arbitrary) order where the predictions for previous labels are included as features for  subsequent models. The resulting rule sets are again purely declarative, but are structured as in 1.\ in the previous section, in that they form a linear order of blocks of rules, one block for each label. Within each block, the order of the rules is interchangable, but each block (semantically) depends on the previous blocks, because of the labels that may appear in its rule bodies. 

One challenge (and opportunity) for research in multi-label rule learning is to break up this rigid block-wise structure into
a more flexible dependency structure. For example, one may consider to adapt the simple covering 
strategy 
from conventional classification to the multi-label case. However, this is not entirely trivial. In the single-label case, this strategy learns one rule at a time, removing all covered examples after a rule has been learned. Thus the next rule is learned only from examples that have not yet been tackled by previous rules. 
A possible adaptation 
is to remove all covered \emph{labels} from these examples, so
that each example remains in the training set until all of its labels
are covered by at least one rule. However, using the resulting rules in a purely declarative setting may be difficult, because if, e.g., one rule $\br_1$ learns to predict label $\lambda_i$, and a subsequent rule $\br_2$ makes use of $\lambda_i$ for predicting $\lambda_j$, then later on, a third rule $\br_3 \neq \br_1$ might be added that also predicts $\lambda_i$, thereby implicitly changing the semantics of $\br_2$. Moreover, $\br_3$ might use $\lambda_j$ or even $\lambda_i$ in its body, leading to cyclic structures that are difficult to handle.
An obvious solution is to treat the resulting rules as a decision list. However, these lists also have to be generalized in that prediction cannot stop after the first rule fires (as in conventional decision lists), but have to be executed in order until the end of the list is reached or a dedicated termination symbol is reached \citep{ML:Rules-Stacking}.

An alternative strategy is to layer the rules not by labels but in a way similar to multi-layer neural networks. Each layer learns single-head rules by a binary relevance algorithm, where the rules in each layer can use the labels that have been predicted in previous layers \cite{ML:Rules-Stacking}. Thus, each layer $i$ introduces labels $h_{i,j}$ corresponding to the labels $\lambda_j$, where the $k$-th layer can make use of all $h_{i,j}$ with $i < k$, and the final layer $K$ is directly mapped to the predictions ($h_{K,j} = \lambda_j$).
This approach, as well as classifier chains, may be viewed as special cases of the 
general framework proposed by Read and Hollm\'en \cite{Read-Hollmen-IDA-14,Read-Hollmen-arxiv-15}, which 
formulates multi-label classification problems as deep networks where label nodes are a special type of hidden nodes which can appear in multiple layers of the networks.


There is ample opportunity for more research on this topic, which tries to combine or generalize these two approaches. For example, Burckhardt and Kramer \cite{Burkhardt2015} generalize BR and CC into 
sequential blocks of sets of labels.
In this setting, each block of rules would again need to have an internal rule to make its prediction. This could be further generalized into a tree-based structure, in which the predictions
of all rules that fire on a path from the root to a leaf are aggregated into a final prediction. Developing a suitable tree-based decision structure for multilabel rules, and an algorithm for deriving this
structure from an already learned list of rules is an open problem.

\section{Constructive Induction}
\label{sec:structure}

For the general case of learning structured declarative rule sets, a key challenge is how to define and train the intermediate concepts $h_i$ which may be used for improving the final prediction. Note that in all examples above, including those for multi-label classification, all $h_i$ always had a fairly clear semantic in that they essentially corresponded to some label $\lambda_j$, and the rule that predicted $h_i$ essentially defined a local pattern for $\lambda_j$ \citep{jf:Dagstuhl-04}. 

However, other hidden concepts may be defined that do not directly correspond to the target variable.
Consider, e.g., the extreme case of learning a parity concept, which checks whether an odd or an even number of $r$ relevant attributes (out of a possibly higher total number of attributes) are set to \emph{true}.
A flat rule-based representation of the target concept for $r = 5$ requires $2^{r-1} = 16$ rules. On the other hand, a layered definition of the predicate, where, e.g., a label $h_1$ is used for defining parity between the first two variables, and $h_i$ uses all $h_{i-1}$ for defining parity between the first $i+1$ variables, would be much more concise, using only $2(r-1) = 8$ rules.

Although there are machine learning systems that can tackle simple problems, 
there is no
system that is powerful enough to learn deeply structured logic theories for realistic problems.
This line of work has been known as
{\em constructive induction\/} \cite{CI-Framework} or \emph{predicate invention}
\cite{ILP-PI},
but surprisingly, it has not received much attention since the classical works in inductive logic programming in the 1980s and 1990s.
One approach is to use a wrapper to scan for regularly co-occurring patterns in rules, and use them to define new intermediate concepts which allow to compress the original theory \cite{AQ17-HCI,CiPF,New-MDL}. 
Alternatively, one can directly
invoke so-called predicate invention operators during the learning process, as, e.g., in Duce \cite{Duce}, which operates in propositional logic, and its successor systems in first-order logic \cite{CIGOL,CHAMP,PI-Statistical}. More recently, \cite{ILP-PI-MetaInterpretative} introduces a technique that employs user-provided meta rules for proposing new predicates, which allow it to invent useful predicates from only very training examples. Kramer \citet{kramer2020} provides an excellent recent summary of work along these lines, pointing out essentially the same research gap as we do in this paper in a much broader context.

A recent attempt to learn a network of rules, in a similar way as in neural networks, revealed the difficulty in inducing useful intermediate concepts when using standard rule learning techniques. In \citet{ba:jung}, several layers of rules were trained by stacking the predictions of the rules of the previous layers, similar to the rule stacking technique of \citet{jf:DS-11}.
By interchanging
layers of conjunctive and disjunctive rules,
these networks are suitable for representing and exploiting auxiliary concepts. However, it turned out that the naive approach of tailoring the rules to directly predicting the target concept and propagating their predictions through subsequent layers prevented
the discovery of complex auxiliary concepts.
This result demands solutions like back-propagation, which learn the target using a backward instead of a forward search, which is subject to future work.

\enlargethispage*{12pt}

\section{Conclusion}
We believe that providing functionalities and support
for learning structured rule bases is crucial for the acceptance of learned models in complex domains. In a way, the recent success of deep neural networks needs to be carried over to the learning of deep logical structures. 
We have briefly touched upon several lines of reserach, which 
may be viewed as a step into this direction, but a clear methodology for learning structured declarative rule sets is still lacking.

\bibliography{bibliography,rules,nn,theory,ml,ilp,ke,jf}
\bibliographystyle{splncs04}

\end{document}